\documentclass[journal]{IEEEtran}
\ifCLASSINFOpdf
\else
\fi

\usepackage{hyperref}
\usepackage{amssymb}
\usepackage{amsmath}
\usepackage{graphicx}
\usepackage{subfigure}
\usepackage{colortbl}
\usepackage{booktabs}

\begin{document}

\title{Faster Image2Video Generation:  A Closer Look at CLIP Image Embedding's Impact on Spatio-Temporal Cross-Attentions}

\author{Ashkan Taghipour,
        Morteza Ghahremani,
        Mohammed Bennamoun,~\IEEEmembership{Senior Member,~IEEE}, \\
        Aref Miri Rekavandi,~\IEEEmembership{Member,~IEEE},
        Zinuo Li,
        Hamid Laga,
        and~Farid Boussaid,~\IEEEmembership{Senior Member,~IEEE}
\thanks{Ashkan Taghipour, Zinuo Li, and Mohammed Bennamoun are with the Department of Computer Science and Software Engineering, The University of Western Australia, Australia. } 
\thanks{Morteza Ghahremani is with the Munich Center for Machine Learning (MCML) and Technical University of Munich (TUM), Germany. }
\thanks{Aref Miri Rekavandi is with The University of Melbourne, Australia. }
\thanks{Hamid Laga is with the School of Information. }
\thanks{Farid Boussaid is with the Department of Electrical, Electronics and Computer Engineering, The University of Western Australia, Australia.}
}

\markboth{Journal of \LaTeX\ Class Files,~Vol.~14, No.~8, August~2015}%
{Shell \MakeLowercase{\textit{et al.}}: Bare Demo of IEEEtran.cls for IEEE Journals}

\maketitle

\begin{abstract}
This paper investigates the role of CLIP image embeddings within the Stable Video Diffusion (SVD) framework, focusing on their impact on video generation quality and computational efficiency. Our findings indicate that CLIP embeddings, while crucial for aesthetic quality, do not significantly contribute towards the subject and background consistency of video outputs. Moreover, the computationally expensive cross-attention mechanism can be effectively replaced by a simpler linear layer. This layer is computed only once at the first diffusion inference step, and its output is then cached and reused throughout the inference process, thereby enhancing efficiency while maintaining high-quality outputs.
Building on these insights, we introduce the VCUT, a training-free approach optimized for efficiency within the SVD architecture. VCUT eliminates temporal cross-attention and replaces spatial cross-attention with a one-time computed linear layer, significantly reducing computational load. The implementation of VCUT leads to a reduction of up to 322T Multiple-Accumulate Operations (MACs) per video and a decrease in model parameters by up to 50M, achieving a 20\% reduction in latency compared to the baseline. Our approach demonstrates that conditioning during the Semantic Binding stage is sufficient, eliminating the need for continuous computation across all inference steps and setting a new standard for efficient video generation. 
\end{abstract}


\section{Introduction}

\noindent
{\small \textit{``Sometimes the most expensive costs are the ones you don't see.''}}\\
{\small \hspace*{\fill}---\textit{Robert Kiyosaki}}
\bigskip % Adds vertical space

\IEEEPARstart{A}dvancements in generative AI have boosted the evolution of innovative technology, initially sparked by generative adversarial networks (GANs)~\cite{tai_gan_image,tai_gan_motion,tai_gan_survey}, across various fields~\cite{molcule_tai} such as image generation \cite{avrahami2024diffuhaul, wang2024stable, taghipour2024box}, motion generation \cite{geng2024text}, robotics~\cite{robotic_tai}, and video generation \cite{ma2024latte}. 
Although there is a growing interest in video modeling, progress is not keeping pace with advancements in image generation and editing. This slower progress can be attributed to several factors, including the high computational demands associated with training on video data \cite{yang2024video, GREEN_TAI}, the scarcity of comprehensive and publicly available video datasets \cite{chen2024panda}, and the complexity inherent in video generation architectures \cite{blattmann2023align, bar2024lumiere, lu2024vdt}.
 Recent developments in diffusion models have garnered considerable attention for their potential in video generation. Building on extensive research in text-to-image (T2I) generation, researchers initially started to explore text-to-video (T2V) models \cite{T2VZero}. These models align video content with corresponding text inputs, producing videos that reflect the provided textual descriptions. However, text-conditioned video generation imposes significant computational burdens, and the alignment between text and video is not optimal \cite{wu2023tune}.
Consequently, recent research has shifted towards image-conditioned video generation. In this approach, a model uses a given initial image, which can be obtained using an off-the-shelf T2I generative model, to generate subsequent video frames. This significantly reduces computational demands and increases generation speed compared to text-based methods. Similar to the T2I generative model, Stability AI \cite{stability_ai} has open-sourced the Stable Video Diffusion (SVD) model \cite{blattmann2023stable} for the image-to-video (I2V) task, positioning it as a pioneering development.

A pivotal component in diffusion architecture that enables the conditioning of generation is the Cross-Attention (CA) mechanism \cite{zhang2024cross}. The CA mechanism plays a crucial role in aligning different modalities, such as text and images, for tasks involving image and video generation. Several studies have explored the importance of CA for spatial control in image generation tasks \cite{rb, Chefer, taghipour2024box}. However, few, if any, have analyzed the role of CA in video generation during the denoising process, particularly from spatial and temporal perspectives.

In this study, we address three new questions regarding the role of CA in the SVD architecture of diffusion models:
\begin{itemize}
    \item \textit{"Is the CLIP image embedding an effective choice for aligning spatial and temporal features in the I2V generation task?"}
    \item \textit{"If necessary, can other computationally efficient architectures achieve the same results as the costly Cross-Attention for conditioning the video generation process?"}
    \item \textit{"Regardless of the architectural selection, be it Cross-Attention or another, is it essential to condition the video generation process at every inference step?"}
\end{itemize}

To investigate these queries, we examine the effectiveness of CLIP image embeddings in capturing spatial and temporal features, and explore the necessity of using the cross-attention mechanism in SVD for conditioning the generated video based on the CLIP image embeddings of the given image within the SVD architecture. Our analysis underscores four important points:

\begin{itemize}
    \item CLIP image embeddings have a negligible impact on the consistency across frames (both subject and background consistency), suggesting that temporal cross-attention within the SVD can be totally discarded without compromising the consistency metrics.
    \item CLIP image embeddings play an important role in the aesthetics and quality of the generated videos; however, this frame quality and aesthetics can be achieved via a simple linear layer, replacing the computation-costly cross-attention mechanism without compromising quality.
    \item There is no need to compute a linear projection of CLIP image embeddings for every step. It can be computed at the first step, cached, and then added to subsequent inference steps.
    \item SVD's inference process primarily consists of two steps, which we call the \textit{Semantic Binding stage} and the \textit{Quality Improvement stage}. Conditioning the video generation during the \textit{Semantic Binding stage} is sufficient, and there is no need to apply Classifier Free Guidance (CFG) \cite{ho2022classifier} at every inference step.

\end{itemize}

These insights led us to develop \textit{Video Computation cUT} (VCUT), a straightforward, effective, and training-free approach designed to enhance the efficiency and preserve the quality of video generation within the Stable Video Diffusion family. Our key observations regarding VCUT include:

\begin{itemize}
    \item VCUT boosts efficiency by eliminating the Temporal Cross Attention (TCA), yet it maintains critical consistency metrics such as subject and background consistency.
    \item VCUT enhances efficiency by re-configuring the SVD architecture. It substitutes the computationally intensive Spatial Cross-Attention (SCA) with a simple linear layer, which is computed only once at the initial inference step, then cached and reused in \textit{semantic binding stages} without any degradation in the aesthetics or image quality of the frames.
    \item The integration of VCUT into the SVD framework can reduce the computational load by up to 322T Multiple-Accumulate Operations (MACs) per video and cut down up to 50M parameters, leading to a 20\% reduction in latency compared to the baseline model, all without incurring additional training costs.
\end{itemize}

In Section \ref{sec:related}, we review related works, followed by a detailed discussion of our proposed method, VCUT, in Section \ref{sec:proposed}. Experimental results are presented in Section \ref{sec:EXP}.

\section{Related Works}
\label{sec:related}

\subsection{Text-to-Video Generation}
Following the success of diffusion models in the T2I generation, research in the field of T2V has gained more attention. A pioneering work in T2V, LVDM \cite{LVDM}, adapted image diffusion models by transforming its 2D UNet architecture into a 3D UNet and trained it on a vast dataset. ModelScope \cite{modelscope}, inspired by LVDM's approach, utilized a more diverse dataset but struggled with handling compositional text descriptions. T2V-Zero \cite{T2VZero}, employing a training-free approach, converted the image diffusion self-attention mechanisms into cross-frame attention; however, its generated motion lacked realism due to the warping method used for frame generation.
VideoFactory \cite{videofactory} introduced a swapped cross-attention mechanism in 3D windows to address temporal distortions, though its performance heavily depends on the distribution of the training data. AnimateDiff \cite{animatediff} incorporated a trained plug-in motion module within the existing T2I architecture, achieving high-quality video generation but failing to optimally align complex texts with the generated videos. Text2Performer \cite{text2performer} focused on the appearance and motion of a human performer to tackle text-video alignment, yet it predominantly generated videos with clean backgrounds and struggled with more complex environments. Lumier \cite{bar2024lumiere} introduced a space-time model capable of generating high-quality videos through a multi-diffusion framework; however, its network parameters are not yet open-sourced, hindering community evaluation of its text-video alignment capabilities. VSTAR \cite{vstar} introduced the concept of temporal nursing in T2V to address the shortcomings of previous models in generating longer videos. Despite tackling longer video generation, it still falls short in optimally aligning videos with complex and lengthy prompts. Following this, Mora \cite{yuan2024mora} proposed to use multiple visual agents for generating longer videos, up to 10 seconds, but it still struggles to encompass all objects mentioned in the text. Consequently, given the existing shortcomings in aligning text with generated videos in T2V models, researchers have explored and tackled similar issues in T2I models \cite{wang2023tokencompose, phung2023grounded}. A new direction suggests using powerful T2I generative models \cite{chen2024pixart} that can optimally align generated images with the provided text. These images could then be used as a basis for video generation, suggesting a shift towards an image-to-video generation approach.

\subsection{Image-to-Video Generation}
VideoCrafter-I2V \cite{chen2023videocrafter1} is a pioneering model in image-to-video (I2V) generation capable of producing videos that adhere to the style, content, and structure of a given reference image. Despite its success, it faces challenges such as unsatisfactory facial representations and inconsistent subject portrayal. ConsistI2V \cite{ren2024consisti2v} proposes a method that enhances subject consistency by utilizing the low-frequency band of the first frame for the noise initialization process. However, it struggles to provide realistic motion, and the videos often exhibit limited motion magnitude, restricting subject movement.
I2V-Adapter \cite{guo2023i2v} preserves the identity of the reference image using cross-frame attention in its architecture. As of this writing, the model's code and checkpoints have not been released, preventing further community assessment of its consistency and motion realism.
SIENE \cite{chen2023seine} attempts to generate long videos at the story level, featuring smooth motion transitions between frames, using an auto-regressive video prediction approach. Although successful in generating story-level videos, it falls short in maintaining background consistency and the aesthetic quality of the frames.
Following this, I2VGen-XL \cite{zhang2023i2vgen} proposes a two-stage video generation process. The first stage, called the base stage, ensures semantic coherency and preserves content, while the second stage, called the refinement stage, enhances details and improves video quality. Despite its high performance, it tends to produce more static frames compared to its competitors, thus limiting the range of subject motions.
DynamicCrafter \cite{xing2023dynamicrafter} addresses the domain limitations of existing video generation models with a dual stream image injection approach that leverages motion priors for open-domain video generation. Despite its success in generating open-domain videos, it faces challenges in maintaining subject consistency when generating videos from complex and highly detailed reference images.

Stable Video Diffusion (SVD) \cite{blattmann2023stable} is an advanced latent video diffusion model designed for high-quality I2V generation. It employs a structured three-phase training process: firstly, image pre-training on the well-known model SD-2\footnote{Stable Diffusion 2: \url{https://huggingface.co/stabilityai/stable-diffusion-2}} to develop robust visual representations; secondly, video pre-training using a large, specially curated video dataset influenced by human preferences; and thirdly, fine-tuning on a select group of high-resolution videos for enhanced quality. This comprehensive approach, enhanced by selective data curation and the integration of temporal layers into the image model, enables SVD to effectively capture dynamic motion and surpass other video generation models in performance. Despite SVD's success in generating high-quality videos with consistent subjects and backgrounds, it is relatively computationally costly. Consequently, in this paper, we introduce VCUT, a method that enhances efficiency. VCUT achieves this by eliminating Temporal Cross Attention and replacing Spatial Cross-Attention with a simple linear layer that is computed once, cached, and reused in \textit{semantic binding stages}, maintaining the aesthetics and quality of the videos. This integration of VCUT into the SVD framework reduces the computational load by up to 322T Multiple-Accumulate Operations (MACs) per video and cuts down up to 50M parameters, which results in a 20\% reduction in latency compared to the baseline model, without the need for additional training costs.

\section{Proposed Method}
\label{sec:proposed}

VCUT is a training-free video generation approach optimized to increase the efficiency of the SVD-based video generation architectures. 
Let $\mathbf{\epsilon}_{\theta} (\mathbf{x}_t,t,y)$, $t\in\{1,\cdots,T\}$, represent a sequence of spatial-temporal denoising UNets with gradients $\nabla{\theta}$  over a batch. 
Sampling in the Latent Diffusion Model (LDM) of video generation employs a decoder to generate a series of RGB frames $\mathbf{x}\in \mathbb{R}^{b \times f\times H\times W\times3}$ from the latent space $\mathbf{z}\in \mathbb{R}^{b \times f\times c\times h\times w}$ that is conditioned on an input image $\mathbf{y}$. Here, $b$ represents the batch size, $f$ denotes the number of frames with height $H$ and width $W$; likewise, $h$, $w$, and $c$ denote the height, width, and the number of channels of the frames in the latent code.
The conditional LDM is trained through $T$ steps:
\begin{align}\label{eq:LDM}
L_{LDM}=\mathop{\mathbb{E}}_{\mathbf{z},\mathbf{y},\mathbf{\epsilon}\sim\mathcal{N}(\mathbf{0}, \mathbf{I})}\left[|| \mathbf{\epsilon} - \mathbf{\epsilon}_{\theta}(\mathbf{z}_t,t,\tau(y))||_2^2
    \right].
\end{align}
In this equation, \(\tau(y)\) is the a pre-trained image embedding that projects the input image \(\mathbf{y}\) into an intermediate representation, and 
\(\theta\) represents the learnable spatial and temporal parameters of the network.

As indicated by Eq. \ref{eq:LDM}, a key step in the I2V generation process is embedding the given image \(\mathbf{y}\) into a space that guides video generation through spatial and temporal cross-attention. The predominant method for image embedding, as used in \cite{wu2024draganything, blattmann2023stable, wang2023motionctrl, ma2024follow, zhang2023i2vgen}, is CLIP image embedding \cite{radford2021learning}.

In this paper, we examine the critical role of this image embedding within the SVD architecture \cite{blattmann2023stable, blattmann2023align} and its impact on network design. Building on the selected CLIP image embedding used in the SVD family, we propose an enhanced and optimized architecture. This new design significantly improves computational efficiency in video diffusion (VD) without compromising video generation quality.

\subsection{CLIP Image Embedding for Temporal Feature Representation% in Video Generation
}
One of the main differences between T2I and I2V generation is the temporal dimension, which plays a crucial role in subject and background consistency. In the SVD framework \cite{blattmann2023stable}, the temporal dimension is addressed by two key blocks:

\begin{itemize}
    \item \textit{TemporalResnetBlock}: Consist of residual blocks based on 3D convolutions \cite{blattmann2023align}.
    \item \textit{TemporalBasicTransformerBlock}: It is designed to ensure subject and background consistency. This block comprises:
    \begin{itemize}
        \item \textit{Temporal Self-Attention (TSA)}: Includes 16 attention mechanisms, including 6 in the encoder, 1 in the middle, and 9 in the decoder section of the denoising \textit{Unet}. It manages intra-frame dependencies to maintain temporal continuity within the video.
        \item \textit{Temporal Cross-Attention (TCA)}: Includes 16 attention mechanisms, with 6 in the encoder, 1 in the middle, and 9 in the decoder section of the denoising \textit{Unet}. It is designed to direct the temporal aspects of generation based on the guided signals (embeddings).

    \end{itemize}
\end{itemize}

To dig deeper into how these temporal attentions perform, consider a 5-D video tensor \( z \), which has dimensions for batch size \( b \), number of latent channels \( c \), number of frames \( f \), height \( h \), and width \( w \). During temporal attention processing, the spatial dimensions (height and width) are merged with the batch dimension. Consequently, the tensor is reshaped into new sequences formatted as \( [b \times h \times w, f, c] \). 
This sequence is then utilized as the query \( Q \), key \( K \), and value \( V \) in the self-attention mechanism is computed as:
\begin{equation} \label{eq:attention}
\text{\textit{TSA}}(z)={\verb|attention|}(Q, K, V) = {\verb|softmax|}\left(\frac{QK^T}{\sqrt{d_k}}\right) V
\end{equation}

\noindent where \( d_k \) represents the scaling factor and based on dimension of $Q$, $K$, and $V$ the the output dimension is $\textit{TSA} \in \mathbb{R}^{(b \times h \times w) \times f \times c}$. 
These self-attention blocks are designed to capture the temporal dynamics within the data, ensuring consistency across frames~\cite{shi2024motion}.

%%%%%%%%%%%%%
\begin{figure*}[t] % 't' positions it at the top of the page
    \centering
    \includegraphics[width=\linewidth]{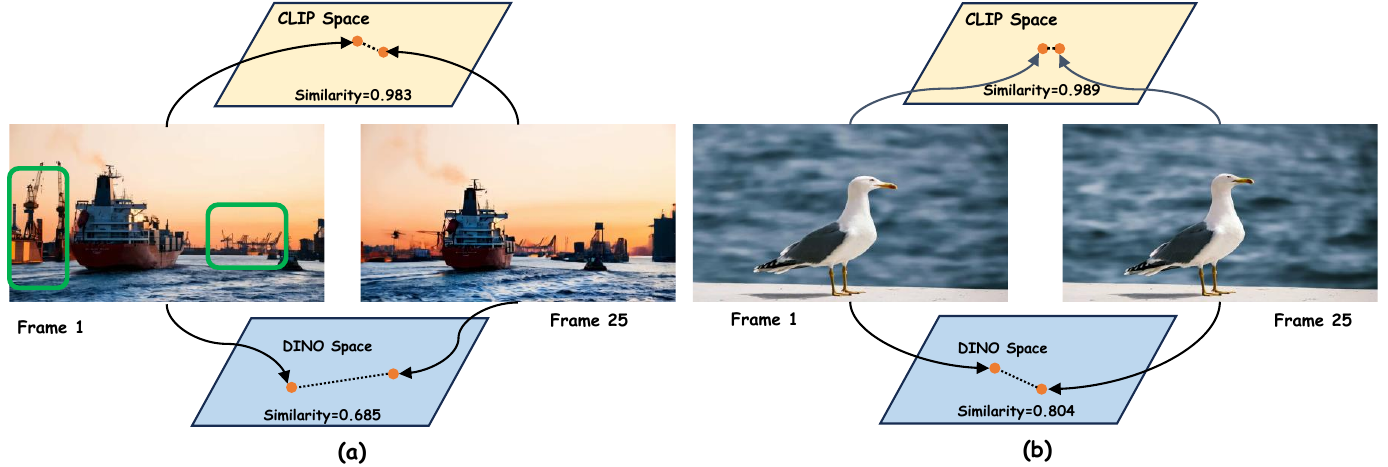}
    \caption{Illustration of the limitation of the CLIP model, which inappropriately assigns high similarity scores despite significant changes in perspective between video frames, suggesting a lack of sensitivity to visual variations. Conversely, the DINO \cite{oquab2024dinov} model more accurately reflects changes, showing lower similarity scores for larger variations.}
    \label{fig:temporalDinoClip}
\end{figure*}
%%%%%%%%%%%%

In \textit{TCA}, \( Q \) is the temporal sequence with shape \( [b \times h \times w, f, c] \). \( K \) and \( V \) are obtained from the CLIP image embedding of the given image, which initially has the shape \( [b, 1, 1024] \). To match dimensions, they are broadcast across the batch dimension, resulting in the shape \( [b \times h \times w, 1, 1024] \). After passing through a linear layer, \( K \) and \( V \) attain the shape \( [b \times h \times w, 1, c] \). Consequently, the attention score \( QK^T \) has the shape \( [b \times h \times w, f, \textbf{1}] \), implying that there is only \textbf{one} key for each query to interact with, which makes the attention score trivial because each frame in \( Q \) attends to only one feature vector in \( K \). Consequently, applying softmax over a dimension of size 1 results in a softmax output of 1, which is redundant. This leads to a constant output after applying the softmax function, indicating that, regardless of \( Q \), the output of the temporal attention score will always be \textbf{one}. This phenomenon renders the cross-attention computation ineffective due to the globally pooled embeddings of CLIP. This leads us to the following insight:

\textit{Insight 1: CLIP image embedding contributes minimally to temporal cross-attention due to its globally pooled nature.} \label{insight:clip-temporal}

\subsection{CLIP Image Embedding for Spatial Feature Representation}
\label{sec:clip_spatial}
In addition to the temporal dimension, the spatial dimension also plays a crucial role in video generation, ensuring the quality of the generated video \cite{Zhou2024storydiffusion}. In SVD architecture \cite{blattmann2023stable}, spatial dimension is addressed by two important blocks:
\begin{itemize}
    \item \textit{ResnetBlock2D}: Primarily applies 2D convolutions \cite{blattmann2023align}  in its ResNet in a way that interprets the video as a batch of independent images.
    \item \textit{BasicTransformerBlock}: Designed to capture the most relevant information in the spatial dimension and guide the video generation in the spatial perspective process. This block comprises:
    \begin{itemize}
        \item \textit{Spatial Self-Attention (SSA)}: which consists of 16 attention mechanisms, including 6 in the encoder, 1 in the middle, and 9 in the decoder section of denoising \textit{Unet}.
        \item \textit{Spatial Cross-Attention (SCA)}: which consists of 16 attention mechanisms applied after \textit{SSA}, including 6 in the encoder, 1 in the middle, and 9 in decoder section of denoising \textit{Unet}.
    \end{itemize}
\end{itemize}

\noindent To further explore the attention mechanism from the spatial perspective, consider the previously defined 5D video tensor $z$  with shape \( [b, c, f, h, w] \). Following the method described by SVD\cite{blattmann2023stable}, the temporal axis is shifted into the batch dimension, interpreting frames as independent images. Consequently, the tensor is reshaped into new sequences formatted as \( [b \times f, h \times w, c] \). This sequence is then used as \( Q \), \( K \), and \( V \) for calculating the \textit{SSA} according to Equation \ref{eq:attention},  resulting in an output tensor of \( \textit{SSA} \in \mathbb{R}^{(b \times f) \times (h \times w) \times c} \).

In \textit{SCA}, \( Q \) is the spatial sequence with shape \( [b \times f, h \times w, c] \). \( K \) and \( V \) are obtained from the CLIP image embedding of the initial image. The shape of this embedding is \( [b, 1, 1024] \). To match dimensions, embeddings are broadcast across the batch dimension, resulting in the shape \( [b \times f, 1, 1024] \). After passing through a linear layer, \( K \) and \( V \) attain the shape \( [b \times f, 1, c] \). Consequently, the attention score \( QK^T \) has the shape \( [b \times f, h \times w, \textbf{1}] \), implying that there is only one key for each query to interact with, which makes the attention score trivial because all pixels in \( Q \) attend to only one feature vector in \( K \). Thus,  similar to \textit{TCA}, regardless of \( Q \), the output of the spatial attention score will always be \textbf{one}, suggesting that the choice of CLIP image embedding makes the cross-attention mechanism trivial and ineffective. This leads us to the second insight of this paper:
% this mechanism can be replaced by a simple linear layer applied to \( V \).

\textit{Insight 2: CLIP image embeddings contribute minimally to spatial cross-attention due to its globally pooled nature.} \label{insight:clip-spatial}

%%%%%%%%%%%%%
\begin{figure*}[t] % 't' positions it at the top of the page
    \centering
    \includegraphics[width=\linewidth]{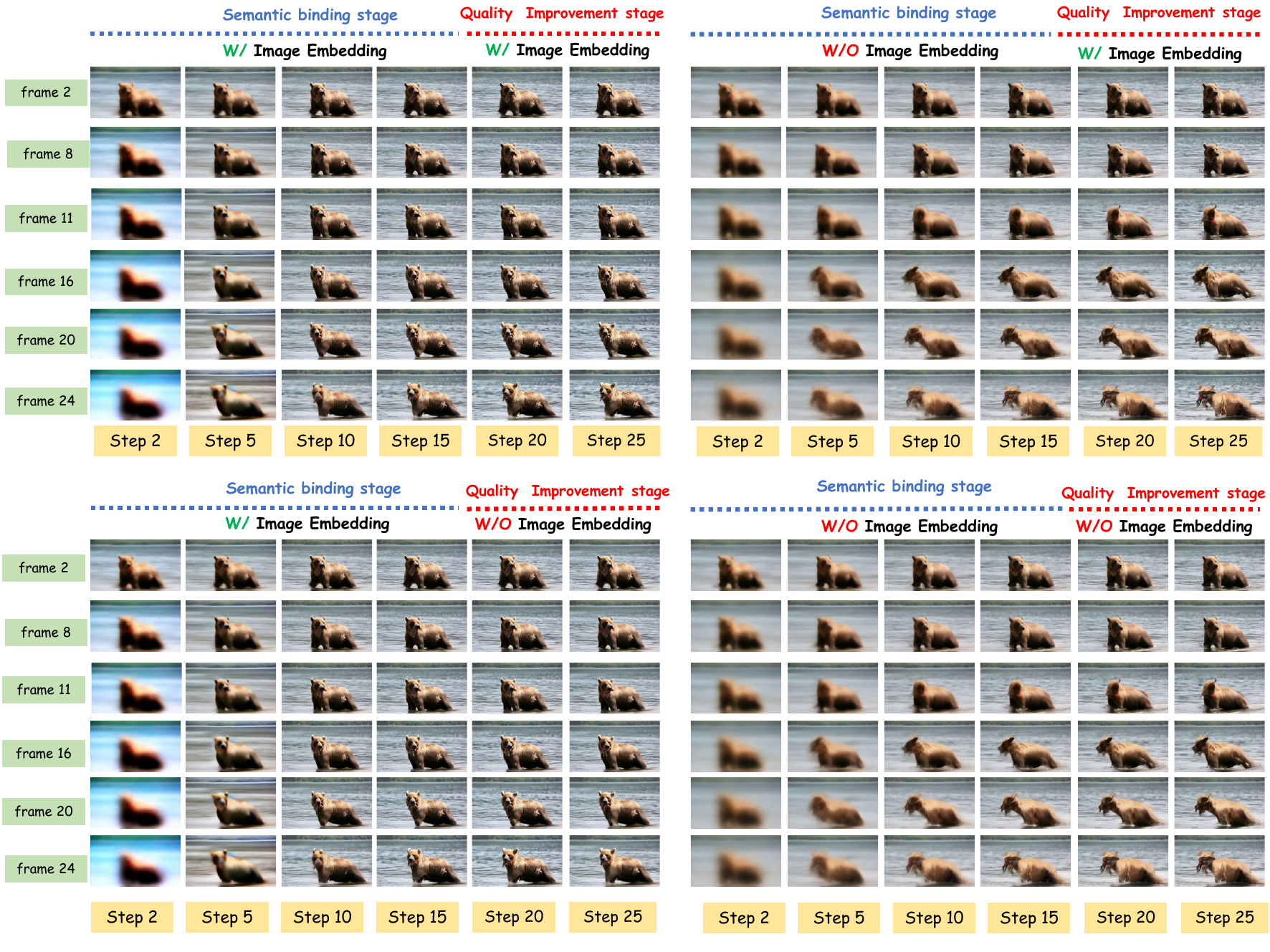}
\caption{This figure illustrates various applications of CLIP image embedding across different stages of the diffusion process. The top left panel shows CLIP image embedding applied at all steps. The top right panel applies it only during the \textit{Quality Improvement stage} (later stages), while the bottom left panel uses it exclusively during the \textit{Semantic Binding stage} (early steps). The bottom right panel does not apply CLIP image embedding at any stage of the diffusion process. This comparison demonstrates that while image embeddings significantly influence the generation process in early stages, their impact lessens in later steps, suggesting that it is feasible to omit embeddings in advanced stages without a loss in image quality.}

    \label{fig:two_stage}
\end{figure*}
%%%%%%%%%%%%

\subsection{Inference Stages In Stable Video Diffusion}
Based on the insights given in \ref{insight:clip-spatial} and \ref{insight:clip-temporal}, the computation of the attention score (\( QK^T \)) is no longer needed because its output invariably consists of a tensor where every element is one. Consequently, the question arises: \textit{Can we completely discard the cross-attention mechanisms?} To address this, one might consider that the outputs of both \textit{TCA} and \textit{SCA} could be replicated by applying a linear layer to the value \( V \). In other words, the functionality achieved by SVD through its cross-attentions in every inference step can be replicated more simply. This is done by applying a linear layer to the CLIP image embedding of the initial frame only once in the first step. The result is then cached and reused in subsequent inference steps.
This approach results in significantly fewer parameters and a lower computational cost.

However, a pertinent question arises: \textit{Can CLIP image embedding capture temporal information, although they were not originally designed for this purpose but based on textual-visual alignment through contrastive loss?} 

To answer this question, we conducted a toy experiment by selecting two videos: one with minimal motion (see Fig.\ref{fig:temporalDinoClip}(b)) and another with considerable camera motion, resulting in significant viewpoint changes (see Fig.\ref{fig:temporalDinoClip}(a)).
We computed the cosine similarity distance between the embeddings of the first and last frames of each video using two different embedding techniques, CLIP \cite{radford2021learning} and DINO \cite{oquab2023dinov2}, as shown in Fig. \ref{fig:temporalDinoClip}.

The experiment revealed very similar cosine similarity scores between the first and last frame embeddings for both videos using CLIP image embedding, as depicted in Fig.\ref{fig:temporalDinoClip}. This outcome suggests that CLIP image embeddings are not very effective at extracting temporal features. Notably, in Fig.\ref{fig:temporalDinoClip}(a), there are objects in the first frame (indicated by green boxes) that do not exist in the last frame. Despite these differences, the cosine similarity score between the first and last frame embeddings of this video is very similar to that of another video that shows negligible motion and has a very similar start and ending frames. 
However, the cosine similarity distance obtained from another image embedding method DINO \cite{oquab2024dinov} shows a considerable difference. Specifically, the cosine similarity distance between the first and last frame embeddings of the first video is much greater than that of the second video. This suggests that DINO provides a more discernible measure of the changes between frames, effectively capturing the temporal differences where CLIP does not.

This toy experiment shows that in addition to problems with granular visual features \cite{tong2024eyes}, CLIP image embeddings also struggle to capture temporal information.
Further experiments, as discussed in Section \ref{sec:EXP}, support this observation, indicating that CLIP image embedding could be entirely discarded from the SVD architecture.
On the other hand, the textual-visual alignment contrastive loss of CLIP image embedding provides a useful representation in the spatial domain. This leads to the conclusion that, instead of discarding the \textit{SCA}, we achieve the same output by replacing \textit{SCA} with a linear layer applied to the CLIP image embedding of the initial frame.
Note that the purpose of this toy experiment is to highlight the shortcomings of CLIP image embedding compared to another image embedding technique. We are not attempting to retrain the video generation network with DINO image embedding, as that is beyond the scope of this paper and not our objective.

\begin{table*}[t]
\centering
\caption{The impact of discarding \textit{TCA} and replacing \textit{SCA} with a linear layer in SVD family architectures on the quality of generated videos across various metrics (\%). "Modified" refers to the architectural changes implemented. Bold values indicate significant improvements in metrics, while non-bold numbers suggest no considerable changes in metrics after \textit{TCA} is removed and \textit{SCA} is replaced with a linear layer.}
\label{tab:effect_of_removing}
\begin{tabular}{@{}lccccccccc@{}}  
\toprule
\toprule
Inference & Subj & BG & Motion & Dynamic & Aesthetics & Imaging & Video-Img. &  Video-Img. \\
Method & Consist & Consist  & Smoothness & Degree & Qual. & Qual. & Subj. Consist. & BG. Consist. \\ 
\midrule
%%%%%%%%
SVD & 96.70 & 96.83 & 97.97 & 43.17 & 60.23& 67.95 & 97.42& 97.68 \\
SVD (Modified) & 96.69 & 96.81 & 97.96 & \textbf{45.44} & 60.17 & 67.63 & 97.40 & 97.67 \\
\midrule
%%%%%%%%%%%%%
SVD-XT & 95.52 & 96.61 & 98.09 & 52.36 & 60.15 & 66.36 & 97.52 & 97.63 \\
% \redrow
SVD-XT (Modified) & 95.60 & 96.46 & 98.03 & \textbf{58.20} & 60.02 & 66.31 & 97.21 & 97.62\\
\midrule
%%%%%%%%%%%%
SVD-XT.1 & 95.42 & 96.77 & 98.12 & 43.17 & 60.23 & 66.78 & 97.51 & 97.62 \\
% \redrow
SVD-XT.1 (Modified) & 95.33 & 96.64 & 98.02 & \textbf{47.55} & 60.11 & 66.50 & 97.50 & 97.65 \\
\bottomrule
\bottomrule
\end{tabular}
\end{table*}

%%%%%%%%%%%%%%%%%%%%%%%%%%%%%%%%%%%%

\subsection{Two Stages of Inference in Stable Video Diffusion}
So far, we have demonstrated that we can discard the \textit{TCA} and replace the \textit{SCA} with a simple linear layer, leading to faster video generation with fewer parameters required. Now, the question arises: \textit{Do we need to apply the classifier-free guidance (CFG) at all diffusion steps, as we do in the SVD architecture and in text-to-image (T2I) generation, considering that we have replaced the cross-attention with a simple linear layer?}

CFG \cite{ho2022classifier} is a method that improves the alignment of generated videos with the initial reference image. It integrates conditional and unconditional generation in \textbf{all} inference steps, enabling improved control over the final output while preserving the quality and diversity of the video \cite{blattmann2023stable}. This process is described by the equation below:
\begin{equation}
    \mathbf{\epsilon}_{\theta}(\mathbf{z}_t,t) = \mathbf{\epsilon}_{\theta}(\mathbf{z}_t,t, \varnothing)+\lambda\left(\mathbf{\epsilon}_{\theta}(\mathbf{z}_t,t,\tau(y))-\mathbf{\epsilon}_{\theta}(\mathbf{z}_t,t, \varnothing)\right)
\end{equation}
 where \(\varnothing\) denotes unconditional generation, a process that generates videos without considering any reference image and its CLIP image embedding.
 In the SVD architecture family, the guidance scale, denoted by \(\lambda\), varies linearly with respect to the number of frames. It starts at 1 and increases linearly to a maximum of 3. Inspired by  \cite{Chefer, llmblue_print, xiao2024rb} which suggested that guiding the diffusion process with directive signals, such as CLIP text embeddings in image generation, is most effective in the early inference stages and less so in later stages, our analysis in the I2V generation task reveals a similar pattern. As illustrated in Fig. \ref{fig:two_stage}, applying CLIP image embedding guidance early on produces the most effective results, whereas its impact diminishes in later stages.

Based on these observations and drawing from practices in image generation, we have divided the video generation inference process into two stages: the \textit{Semantic Binding stage} and the \textit{Quality Improvement stage}. The first stage directs the video generation process to align semantically with the reference image, leveraging the effectiveness of CLIP image embeddings. The second stage, termed the \textit{Quality Improvement stage}, primarily focuses on the denoising process of diffusion models, where CLIP image embeddings prove to be less effective.

Consequently, within this two-stage framework, we propose replacing the output of two cross-attention maps, $SCA_{\tau(y)}^{c,m}$ and $SCA_{\varnothing}^{c,m}$, with outputs from two simpler linear layers, $L_{\tau(y)}^{c,m}$ and $L_{\varnothing}^{c,m}$, starting from a specific step \(c\) onward. This is defined mathematically as:

\begin{equation}
M = \left\{ \frac{1}{2} (L_{\tau(y)}^{c,m} + L_{\varnothing}^{c,m}) \mid m \in [1, l] \right\}
\end{equation}

where $L_{\tau(y)}^{c,m}$ denotes the linear layer applied to the CLIP image embedding of the reference image, $L_{\varnothing}^{c,m}$ represents the output from the linear layer for unconditional generation (similar to CFG), and $l$ is the total number of 16 linear layers (previously SCA) applied to the CLIP image embedding of the reference image. Applying VCUT from the specific step $c$ onward means that guidance is only applied during the \textit{Semantic Binding stage}. The differences in the \(c\) cut steps will be analyzed in the experimental results section.

 %%%%%%%
\begin{table}[t]
\centering
\caption{Computation Cost Savings from Removing  Temporal Cross Attention and Replacing Spatial Cross Attention with a Linear Layer on MACs and Params. The $\downarrow$ denotes the amount of decrease in each metric.}

\label{tab:single_mac}
\begin{tabular}{@{}lcccc@{}}
\toprule
\toprule
Inference Method & \textit{TCA} & \textit{SCA}& MACs (T) & Params. (B) \\ \midrule
SVD & \checkmark  & \checkmark  &36.11 & 1.521   \\ 
SVD & \(\times\) & \(\times\) &35.1$_{1T\downarrow}$ & 1.474$_{47M\downarrow}$ \\
\midrule
SVD-XT &  \checkmark  & \checkmark  &64.41 & 1.524  \\
SVD-XT &  \(\times\) & \(\times\) &62.86$_{1.5T\downarrow}$ & 1.474$_{50M\downarrow}$  \\
\midrule
SVD-XT.1 &  \checkmark  & \checkmark  &64.41 & 1.524  \\
SVD-XT.1 &  \(\times\) & \(\times\) &62.86$_{1.5T \downarrow}$ & 1.474$_{50M\downarrow}$\\
\bottomrule
\bottomrule
\end{tabular}
\end{table}
%%%%%

\section{Experiments}
\label{sec:EXP}
 
In this section, we evaluate the performance of the proposed method from two distinct perspectives to comprehensively assess its efficacy. Firstly, we consider the computational burden of the method, focusing on its efficiency in reducing computation operations. Secondly, we examine the quality of the generated videos. For each of these perspectives, specific metrics are employed to quantitatively measure performance. We will explore the detailed evaluation metrics used for each perspective, providing a structured framework for our analysis.

%%%%%%%
\begin{table}[t]
\centering
\caption{Computation Cost Savings from Removing TCA, Replacing SCA with a Linear Layer, and Integrating the VCUT Technique at Various Steps into the SVD Family: Impact on MACs, Params, and Latency. The $\downarrow$ denotes a reduction in each metric.}

\label{tab:total_save_macs}
\begin{tabular}{@{}lcccc@{}}
\toprule
\toprule
Inference Method & MACs & Params.  & Latency 
\\ & (T) & (B) & (s)
\\
\midrule

SVD & 903 & 1.521 & 68.4 \\ 
SVD+VCUT (c=17) & 719$_{184T\downarrow}$ & 1.474$_{47M\downarrow}$& 54.7$_{20\% \downarrow}$  \\
SVD+VCUT (c=20) & 772$_{131T\downarrow}$ & 1.474$_{47M\downarrow}$ & 58.2$_{15\% \downarrow}$  \\
\midrule
SVD-XT & 1610 & 1.521 & 120.6  \\ 
SVD-XT+VCUT (c=17) & 1288$_{322T\downarrow}$& 1.474$_{50M\downarrow}$ & 97.3$_{19\%\downarrow}$\\ 
SVD-XT+VCUT (c=20) & 1382$_{228T\downarrow}$ & 1.474$_{50M\downarrow}$ & 103.2$_{14\% \downarrow}$ \\ 
\midrule
SVD-XT.1 & 1610 & 1.524 & 119.8  \\
SVD-XT.1+VCUT (c=17) & 1288$_{322T\downarrow}$ & 1.474$_{50M\downarrow}$  & 97.1$_{19\% \downarrow}$\\ 
SVD-XT.1+VCUT (c=20) & 1382$_{228T\downarrow}$ & 1.474$_{50M\downarrow}$  & 102.8$_{14\%\downarrow}$\\ 
\bottomrule
\bottomrule
\end{tabular}
\end{table}
%%%%%

\subsection{Evaluation of Computational Efficiency}

To assess the computational efficiency of our proposed method, we utilize several key metrics. Firstly, we count the Multiple-Accumulate Operations (MACs) \cite{pytorchOpCounter}, which are crucial for understanding computational complexity. Additionally, we evaluate the total number of parameters (Params.) in our model to determine the impact of proposed changes in terms of a reduction in the number of parameters which is one of the vital criteria in modern computer vision systems \cite{rekavandi2023transformers,rekavandi2022guide}. We measure the latency per sample to gauge the time savings achieved by integrating our method into the network. This latency measurement is conducted on an NVIDIA A10 graphics card with 24GB VRAM.

\subsection{Evaluation of Generated Videos}
To precisely assess the effectiveness of the proposed approach,  we use the VBench \cite{huang2023vbench} video generation assessment suite, which focuses on various temporal and spatial dimensions,

\subsubsection{Temporal Analysis}
For the temporal assessment, we use the bellow metrics:
\begin{itemize}
    \item Subject Consistency: It measures the identity variations of the subject within the video. It employs DINO~\cite{oquab2024dinov,oquab2023dinov2} feature extraction method to compute distances between features extracted from the subject in the \textbf{generated} frames, as described by the following equation:
    \begin{equation}\label{eqsubj_cons}
        S_{\text{subject}} = \frac{1}{T-1} \sum_{t=2}^{T} \frac{1}{2} \left( \langle d_1, d_t \rangle + \langle d_{t-1}, d_t \rangle \right) 
    \end{equation}
    where \(d_t\) denotes the normalized DINO image feature of the \(t\)-th frame, \(T\) is the number of frames, and \(\langle \cdot, \cdot \rangle\) represents the dot product for cosine similarity calculations.

    \item Video-Image Subject Consistency: It assesses the identity consistency between the subject in the \textbf{reference} image and subsequent frames, using cosine similarity calculations of DINO features as specified:
    \begin{equation}\label{eq:i2v_subj-cons}
        S_{\text{VI-Subj-cons}} = \frac{1}{T-1} \sum_{t=2}^{T} \frac{1}{2} \left( \langle d_r, d_t \rangle + \langle d_{t-1}, d_t \rangle \right) 
    \end{equation}
    where \(d_r\) is the normalized DINO image feature of the reference image, aligning with the computation and definitions in Eq. \ref{eqsubj_cons}.

    \item Background Consistency: This metric evaluates the uniformity of the background across different frames to ensure visual continuity. It is calculated using the CLIP image encoder \cite{radford2021learning} to extract feature vectors:
    \begin{equation}\label{eq:bg-cons}
        S_{\text{BG-Consist}} = \frac{1}{T-1} \sum_{t=2}^{T} \frac{1}{2} \left( \langle c_1, c_t \rangle + \langle c_{t-1}, c_t \rangle \right)
    \end{equation}
    where \(c_i\) represents the normalized CLIP image feature of the \(i\)-th frame.

    \item Video-Image Background Consistency: It assesses the background consistency between the reference image and subsequent frames using cosine similarity calculations:
    \begin{equation}\label{eq:i2v_bg-cons}
        S_{\text{VI-BG-Consist}} = \frac{1}{T-1} \sum_{t=2}^{T} \frac{1}{2} \left( \langle c_r, c_t \rangle + \langle c_{t-1}, c_t \rangle \right)
    \end{equation}
    where \(c_r\) is the normalized CLIP image feature of the reference image, consistent with the parameters in Eq. \ref{eq:bg-cons}.

    \item Motion Smoothness: It measures whether the motion in the generated video is smooth and follows the physical law. Specifically, for a video with frames \([f_0, f_1, f_2, f_3, \ldots, f_{2n-2}, f_{2n-1}, f_{2n}]\), the odd-numbered frames are removed and video frame interpolation \cite{licvpr23amt} is applied to estimate the missing frames \([f_1, f_3, \ldots, f_{2n-1}]\). The \textit{MAE} is then calculated between these interpolated frames and the original odd-numbered frames. This MAE is normalized resulting in a final score ranging from \([0, 1]\), where a higher score denotes greater smoothness of motion.

     \item Dynamic Degree: It measures how a model tends to generate static videos that result in a lack of motion. To calculate this, the average of the largest 5\% of RAFT \cite{teed2020raft} optical flows is considered as the basis. The final Dynamic Degree score is calculated by measuring the proportion of non-static videos generated by the model.

\end{itemize}

\begin{table*}[t]
\centering
\caption{Quantitative Analysis of Integrating the Proposed VCUT into the SVD Family Models at Different Cut Steps ``c"}

\label{tab:different_steps_quality}
\begin{tabular}{@{}lccccccccc@{}} 
\toprule
\toprule
Inference & Subj. & BG & Motion & Dynamic & Aesthetics & Imaging & Video-Img. &  Video-Img. & Latency \\
Method & Consist. & Consist.  & Smoothness   & Degree & Qual.   & Qual.   & Subj. Consist.  & BG. Consist.    &  (s) \\  
&(\%) &(\%) &(\%) &(\%) &(\%) &(\%) &(\%) &(\%) &\\
\midrule
%%%%%%%%
SVD & 96.70 & 96.83 & 97.97 & 43.17 & 60.23 & 67.95 & 97.42 & 97.68 & 68.4\\
SVD\textbf{+}VCUT (c=10) & 91.85 & 91.23 & 94.07 & 43.20 & 52.34 & 61.42 & 91.09 & 91.88 & 47.1\textcolor{red}{$_{31\% \downarrow}$}\\
SVD\textbf{+}VCUT (c=17) & 96.69 & 96.88 & 98.04 & \textbf{45.77} & 60.15 & 67.46 & 97.32 & 97.49 & 54.7\textcolor{red}{$_{20\% \downarrow}$}\\
SVD\textbf{+}VCUT (c=20) & 96.70 & 96.91 & 97.92 & \textbf{44.05} & 60.19 & 67.67 & 97.42 & 97.60 & 58.2\textcolor{red}{$_{15\% \downarrow}$}\\
\midrule
%%%%%%%%%%%%%
% \redrow
SVD-XT & 95.52& 96.61 & 98.09 & 52.36 & 60.15 & 66.36 & 97.52 & 97.63 & 120.6\\
SVD-XT\textbf{+}VCUT (c=10) & 91.11 & 90.02 & 94.81 & 54.27 & 53.94 & 60.53 & 91.86 & 92.19 & 84.3\textcolor{red}{$_{30\% \downarrow}$}\\
SVD-XT\textbf{+}VCUT (c=17) & 95.48 & 96.49 & 98.04 & \textbf{57.80} & 60.07 & 66.11 & 97.25 & 97.52 & 97.3\textcolor{red}{$_{19\% \downarrow}$}\\
SVD-XT\textbf{+}VCUT (c=20) & 95.54 & 96.53 & 98.03 & \textbf{58.69} & 60.10 & 66.17 & 97.42 & 97.61 & 103.2\textcolor{red}{$_{14\% \downarrow}$}\\
\midrule
%%%%%%%%%%%%
SVD-XT.1 & 95.42 & 96.77 & 98.12 & 43.17 & 60.23 & 66.78 & 97.51 & 97.62 & 119.8\\
SVD-XT.1\textbf{+}VCUT (c=10) & 91.22 & 91.14 & 94.97 & 50.09 & 54.28 & 59.58 & 92.23 & 92.12 & 83.8\textcolor{red}{$_{30\% \downarrow}$}\\
SVD-XT.1\textbf{+}VCUT (c=17) & 95.33 & 96.64 & 98.04 & \textbf{52.76} & 60.16 & 66.41 & 97.23 & 97.51 & 97.1\textcolor{red}{$_{19\% \downarrow}$}\\
SVD-XT.1\textbf{+}VCUT (c=20) & 95.37 & 96.69 & 98.06 & \textbf{52.52} & 60.19 & 66.62 & 97.35 & 97.58 & 102.8\textcolor{red}{$_{14\% \downarrow}$}\\
\bottomrule
\bottomrule
\end{tabular}
\end{table*}

\subsubsection{Spatial Analysis}

\begin{itemize}
    
    \item Aesthetic Quality: measures photographic composition, color richness, and artistic subject quality using the LAION aesthetic predictor \cite{laion}. Each frame is rated from 0 to 10, and these scores are linearly normalized to a 0-1 scale. The average of these normalized scores determines the video's overall aesthetic score.
    
    \item Imaging Quality: measures distortions such as over-exposure, noise, and blur in the generated frames. This is assessed using the MUSIQ \cite{ke2021musiq} image quality predictor, which is trained on the SPAQ \cite{SPAQ} dataset.

\end{itemize}

\subsection{Base Models}
We utilize all three released models from the SVD family as the baseline, including SVD\footnote{\url{https://huggingface.co/stabilityai/stable-video-diffusion-img2vid}}, SVD-XT\footnote{\url{https://huggingface.co/stabilityai/stable-video-diffusion-img2vid-xt}}, and SVD-XT.1\footnote{\url{https://huggingface.co/stabilityai/stable-video-diffusion-img2vid-xt-1-1}}. The first model, SVD, is trained to generate video clips from an image condition, producing 14 frames at a resolution of 576$\times$1024. The second model, SVD-XT, also generates videos at this resolution but increases the frame count to 25. This model was fine-tuned from the original SVD. The third model, SVD-XT.1, designed to generate 25 frames at a resolution of 576$\times$1024, was fine-tuned from SVD-XT to enhance the consistency and quality of the outputs.

\subsection{Image Benchmark Suit}
To generate videos from images, we utilize the Vbench \cite{huang2023vbench} image suite, which features a diverse and equitable selection of content for both foreground and background elements. This ensures variability in several aspects: scene category, object type, and fairness in human-centric images. The Vbench image suite comprises a total of 1,118 images. For benchmarking purposes, in accordance with Vbench guidelines, five videos are generated from each image, each with different random noise. These random seeds will be kept consistent across various setups to ensure fair comparisons.
 This process results in a total of 5,590 generated videos. The evaluation metrics are computed based on this setup.

\subsection{Impact of Removing TCA and Replacing SCA with Linear Layer}
As demonstrated in Fig. \ref{fig:temporalDinoClip} through a toy experiment, CLIP Image embeddings are not ideal for preserving temporal information in video frames. To quantitatively assess the impact of removing the \textit{TCA} and replacing the \textit{SCA} with a linear layer on the video generation process, we modified SVD family models accordingly and evaluated the generated videos using the Vbench image benchmark suite \cite{huang2023vbench} from various aspects, including temporal consistency, motion capability, and imaging quality.
The results, presented in Table \ref{tab:effect_of_removing}, show that the performance metrics do not significantly decline, indicating the ineffectiveness of \textit{TCA} for temporal consistency (both \textit{subject} and \textit{background}). Furthermore, this suggests that \textit{SCA} can be replaced with a simple linear layer without compromising the aesthetics or image quality of the generated videos. 

Moreover, the results show that the proposed changes lead to an increase in the dynamic degree metric, which we believe allows the generated video to exhibit more motion than the original architecture. This enhancement occurs because the modified approach removes the temporal restrictions imposed by the TCA on each frame's adherence to the CLIP image embedding of the reference image. Originally, this adherence did not improve the consistency metric and actually diminished the dynamic degree.

%%%%%%%%%%%%%
\begin{figure*}[t] 
    \centering
    \includegraphics[width=\linewidth]{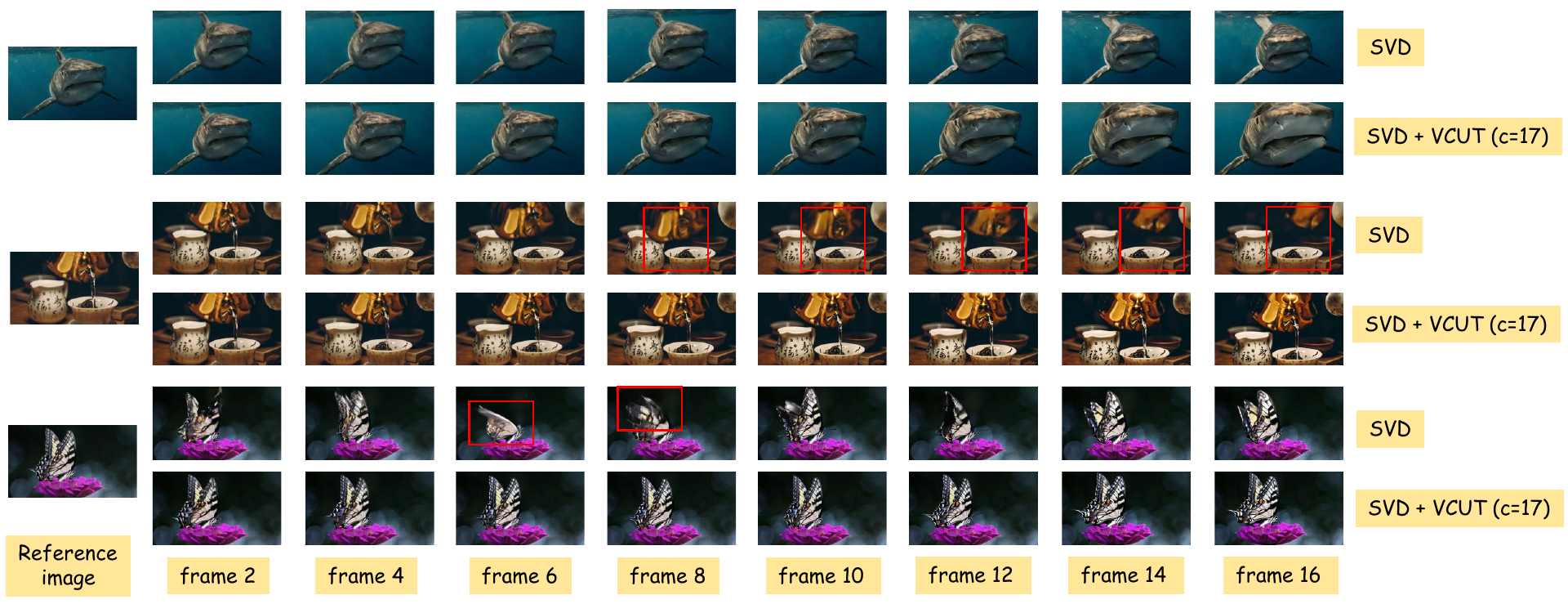}
    \caption{The figure showcases video frames generated by standard SVD and VCUT-integrated SVD models. The top two rows demonstrate enhanced motion in the \textit{shark} video, highlighting the benefits of VCUT in dynamic degree metric as noted in Table \ref{tab:different_steps_quality}. Middle rows show that removing \textit{TCA} improves the consistency of \textit{teapot} videos, indicated by red boxes. The bottom rows confirm that VCUT does not compromise spatial quality, as seen in the less blurry butterfly frames. These comparisons illustrate VCUT's effectiveness in enhancing video dynamics and quality without additional computational costs.}
    \label{fig:visual_compare}
\end{figure*}
%%%%%%%%%%%%

\subsection{Computation Complexity Improvement over Base Models}
We implemented the proposed modifications in cross-attention across all three models of the SVD family. As demonstrated in Table \ref{tab:single_mac}, discarding \textit{TCA} and replacing \textit{SCA} with a simpler linear layer leads to significant reductions at \textbf{each} diffusion step—up to 1.5 trillion MACs and millions of parameters. This reduction is crucial, as the diffusion process involves multiple denoising steps, cumulatively enhancing computational efficiency throughout the video generation process.

Building on these findings, Table \ref{tab:total_save_macs} quantifies the overall impact of these changes across the entire SVD family. By integrating the removal of \textit{TCA}, replacing \textit{SCA} with a linear layer, and applying the VCUT approach at various inference steps, we achieve substantial computational savings. Specifically, these modifications reduce up to 322T MACs and decrease up to 50M parameters, leading to an approximate 20\% reduction in inference time. This represents a time savings of 35 hours when generating 5590 videos using the Vbench\cite{huang2023vbench} image suite. Notably, these computational efficiencies are achieved without compromising the quality of the generated videos—a critical aspect that we will explore in further detail later in \ref{subsec:Quality_preserve}.

\subsection{Quality preservation of the Base Models}
\label{subsec:Quality_preserve}
In Table \ref{tab:different_steps_quality}, we present a quantitative analysis of integrating the proposed VCUT method into the SVD family at various empirical cut steps \(c = 10, 17,\) and \(20\). The results show that applying the VCUT technique at early steps, such as \(c = 10\), significantly decreases the computation burden (\textit{Latency}) up to 31\%, 30\%, and 30\% for SVD, SVD-XT, and SVD-XT.1 correspondingly, but at the cost of compromising the consistency-related metrics (\textit{subject and background consistency}) and video quality (\textit{Aesthetics and Imaging Quality}). This shows that in the \textbf{Semantic Binding stages} of the video generation process, utilizing the CLIP image embedding and CFG is necessary to ensure the quality of generated videos.

However, integrating the VCUT technique at later steps (\(c = 17\) or \(20\)) ensures that neither the consistency metrics (\textit{subject and background consistency}) nor the quality metrics (\textit{Aesthetics and Imaging Quality}) of the generated videos are compromised, and simultaneously boosts the generation speed up to 20\%, 19\%, and 19\% for SVD, SVD-XT, and SVD-XT.1, respectively.

It is important to note that integrating the VCUT at \(c = 17\) and \(c = 20\) steps also leads to higher \textit{Dynamic Degree} metrics, indicating the production of more dynamic videos. This is particularly relevant as many existing video generation frameworks primarily produce highly static frames \cite{zhang2023i2vgen, chen2023seine}, which can misleadingly inflate consistency-related scores. These frameworks often fail to achieve acceptable motion in frames, a challenge that the VCUT integration addresses by enhancing video dynamism.

In Fig. \ref{fig:visual_compare}, we present samples from videos generated using the standard SVD and the VCUT-integrated versions. The first two rows show a \textit{shark} example where VCUT integration enhances motion towards the camera, illustrating increased dynamism and justifying the improved dynamic degree metric shown in Table \ref{tab:different_steps_quality}. The next two rows, highlighted by red boxes, demonstrate that VCUT does not compromise but in some cases rather enhances the consistency of the generated \textit{teapot} video. This suggests that removing \textit{TCA} does not negatively impact, and can even improve, consistency by reducing the restrictions imposed by alignment with CLIP image embeddings, which predominantly preserve global rather than local fine-grained features.

The final two rows of Fig. \ref{fig:visual_compare} show that integrating VCUT with the base SVD network does not degrade, and can enhance, the spatial quality of the generated frames. In some cases, indicated by red boxes, the butterfly appears less blurry, demonstrating an improvement in frame quality.

\section{Conclusion}
In this paper, we explore the role of the Cross-Attention mechanism within the Stable Video Diffusion (SVD) family and its application in guiding the video generation process based on CLIP image embeddings. We demonstrate that neither temporal nor spatial cross-attention is essential within the SVD framework. Temporal cross-attention can be entirely discarded, and spatial cross-attention can be replaced with a simple linear layer without sacrificing the consistency and quality metrics of the generated videos. Based on these insights and an analysis of how granularly CLIP image embeddings preserve features of the input image, we propose the VCUT method. VCUT is a training-free approach that can be integrated into the SVD family during inference to accelerate video generation without compromising quality.

% \section*{Acknowledgment}
% We would like to acknowledge the use of GPT-4 for proofreading and spellchecking this paper.

\ifCLASSOPTIONcaptionsoff
  \newpage
\fi

\bibliographystyle{IEEEtran}
\bibliography{main}

% \end{IEEEbiography}


% % insert where needed to balance the two columns on the last page with
% %\newpage



%\vfill


\end{document}